\pdfoutput=1

\documentclass[11pt]{article}

\usepackage{acl}

\usepackage{times}
\usepackage{latexsym}
\usepackage{graphics}
\usepackage{graphicx}
\usepackage{color}
\usepackage{colortbl}
\usepackage{amssymb}
\usepackage{amsmath}
\usepackage{bbm}
\usepackage{booktabs}

\usepackage[T1]{fontenc}

\usepackage[utf8]{inputenc}

\usepackage{microtype}
\usepackage{multirow}
\usepackage{footmisc}
\usepackage{soul}
\usepackage{arydshln}
\usepackage{CJKutf8}

\title{Rethink about the Word-level Quality Estimation for Machine Translation from Human Judgement}

\author{Zhen Yang,  Fandong Meng, Yuanmeng Yan, and Jie Zhou \\
   Pattern Recognition Center, WeChat AI, Tencent Inc, Beijing, China \\
  {\tt \{zieenyang, fandongmeng, withtomzhou\}@tencent.com}}
  
\begin{document}
\begin{CJK}{UTF8}{gbsn}
\maketitle

\begin{abstract}
Word-level Quality Estimation (QE) of Machine Translation (MT) aims to find out potential translation errors in the translated sentence without reference. Typically, conventional works on word-level QE are designed to predict the  translation quality in terms of the post-editing effort, where the word labels ("OK" and "BAD") are automatically generated by comparing words between MT sentences and the post-edited sentences through a Translation Error Rate (TER) toolkit. While the post-editing effort can be used to measure the translation quality to some extent, we find it usually conflicts with the human judgement on whether the word is well or poorly translated.  To overcome the limitation, we first create a golden benchmark dataset, namely \emph{HJQE} (Human Judgement on Quality Estimation), where the expert translators directly annotate the poorly translated words on their judgements. Additionally, to further make use of the parallel corpus, we propose the self-supervised pre-training with two tag correcting strategies, namely tag refinement strategy and tree-based annotation strategy, to make the TER-based artificial QE corpus closer to \emph{HJQE}. We conduct substantial experiments based on the publicly available WMT En-De and En-Zh corpora. The results not only show our proposed dataset is more consistent with human judgment but also confirm the effectiveness of the proposed tag correcting strategies.\footnote{The data can be found at \url{https://github.com/ZhenYangIACAS/HJQE}.}
\end{abstract}

\section{Introduction}
\begin{figure}
    \centering
    \resizebox{.48\textwidth}{!}{
    \includegraphics{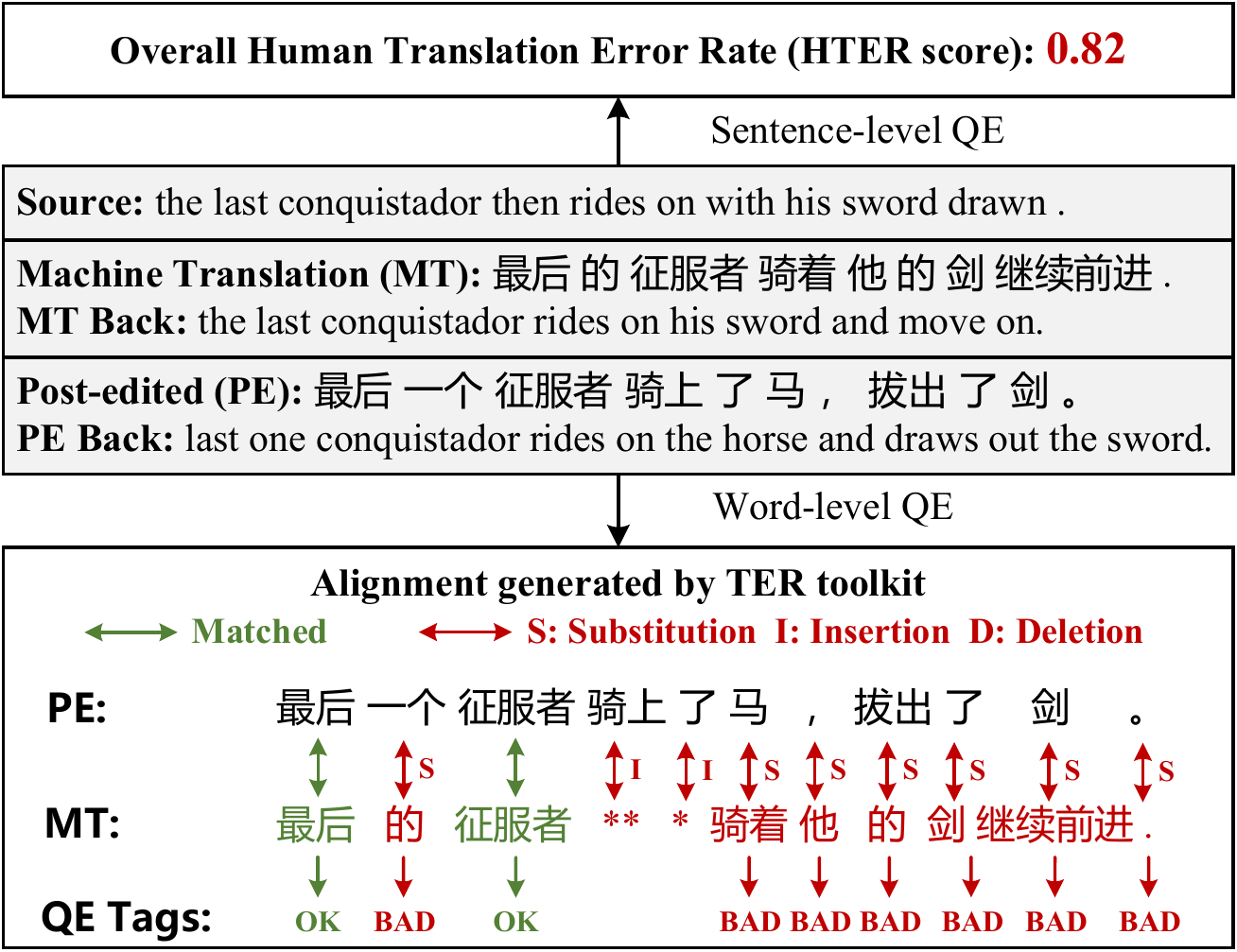}}
    \caption{The illustration of the sentence-level and word-level QE tasks. The word-level QE tags are generated by the TER toolkit.}
    \label{fig:intro1}
    \vspace{-0.2cm}
\end{figure}

Quality Estimation (QE) of Machine Translation (MT) aims to automatically estimate the quality of the translation generated by MT systems, with no reference available. It typically acts as a post-processing module in commercial MT systems, alerting the user with potential translation errors. Figure \ref{fig:intro1} shows an example of QE, where the sentence-level task predicts a score indicating the overall translation quality, and the word-level QE needs to predict each word as \texttt{OK} or \texttt{BAD}\footnote{In this paper, we mainly focus on the word-level QE on the target side, which aims to detect potential translation errors in MT sentences.}.

\begin{figure*}
    \centering
    \resizebox{.84\textwidth}{!}{
    \includegraphics{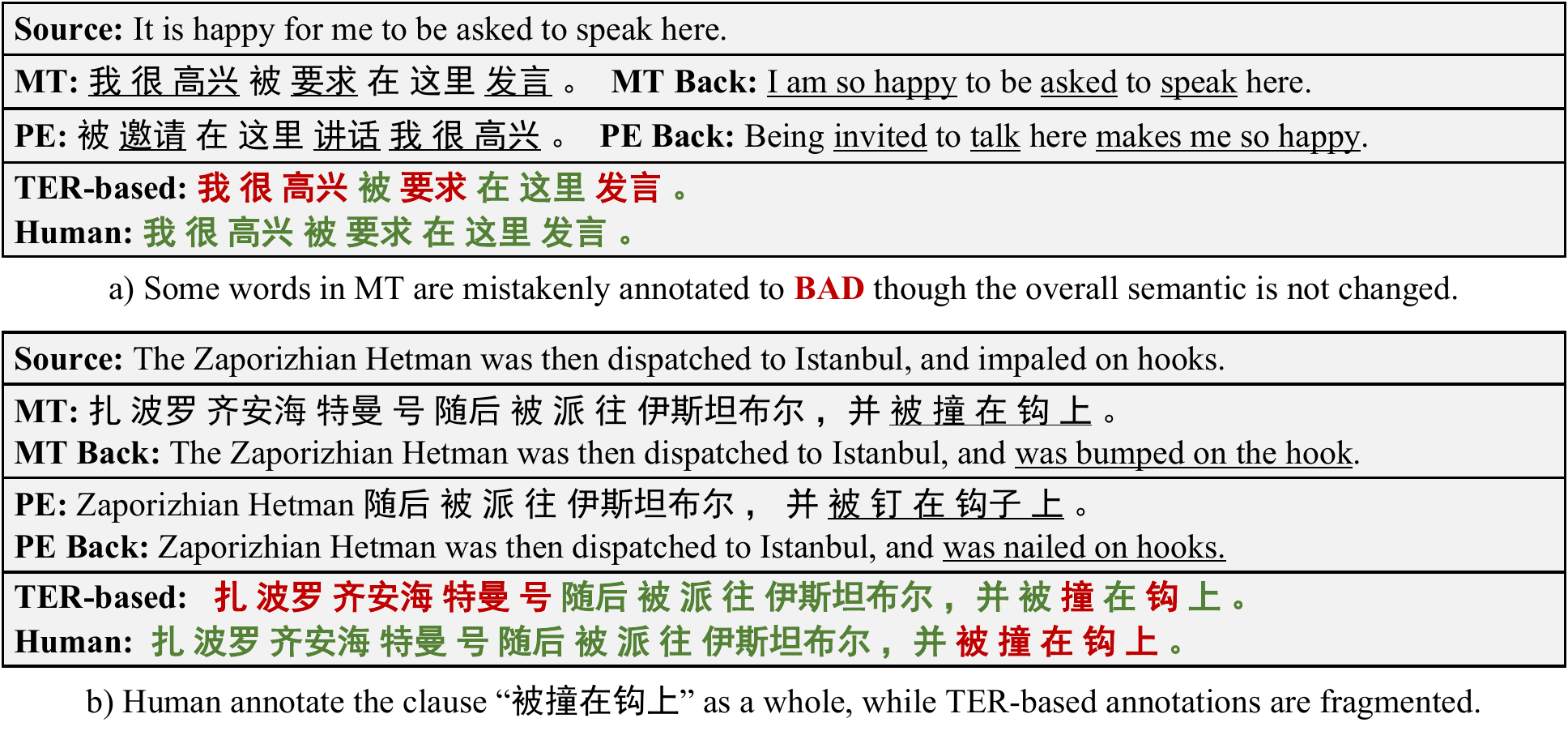}}
    \caption{Two examples show the gap between the TER-based annotation and human's direct annotation on detecting translation errors. The red color indicates \texttt{BAD} tags (text with translation errors), while the green color indicates \texttt{OK} tags (text without translation errors).}
    \label{fig:intro2}
    \vspace{-0.2cm}
\end{figure*}

Currently, the collection of QE datasets mainly relies on the Translation Error Rate (TER) toolkit \cite{snover2006study}. Specifically, given the machine translations and their corresponding post-edits (PE, generated by human translators) or target sentences of parallel corpus as the pseudo-PE \cite{tuan-etal-2021-quality, lee2020two}, the rule-based TER toolkit is used to generate the word-level alignment between the MT and the PE based on the principle of minimal editing. All MT words not aligned to PE are annotated as \texttt{BAD} (shown in Figure \ref{fig:intro1}). Such annotation is also referred as post-editing effort \cite{fomicheva2020mlqe, specia-etal-2020-findings-wmt}. The post-editing effort measures the translation quality in terms of the efforts the translator need to spend to transform the MT sentence to the golden reference. However, we find it usually conflicts with human judgements on whether the word is well or poorly translated. There are two main issues that result in the conflicts. First, the PE sentences often substitute some words with better synonyms and reorder some sentence constituents for polish purposes. These operations do not destroy the meaning of the translated sentence, but make some words mistakenly annotated under the exact matching criterion of TER (shown in Figure \ref{fig:intro2}a). Second, when fatal errors occur in MTs, a human annotator typically takes the whole sentence or clause as \texttt{BAD}. However, the TER-based annotation still try to find trivial words that align with PE, resulting in fragmented annotations (shown in Figure \ref{fig:intro2}b). In many application scenarios and down-stream tasks, it is usually important even necessary to detect whether the word is well or poorly translated from the human judgement \cite{yang2021wets}. However, most previous works still use the TER-based dataset as the evaluation benchmark, which makes the models' predictions deviate from the human judgement.

To overcome the limitations stated above, for the first time, we concentrate on the model's ability of finding translation errors on MT sentences from the human judgement. We first collect a high quality benchmark dataset \emph{HJQE} where human annotators directly annotate the text spans that lead to the translation errors in MT sentences. Then, to further make use of the large scale translation parallel corpus, we also propose two tag correcting strategies, namely tag refinement strategy and tree-based annotation strategy, which make the TER-based annotations more consistent with human judgment.


Our contributions can be summarized as follows: 1) We collect a new dataset called \emph{HJQE} that directly annotates translation errors on MT sentences. We conduct detailed analyses and demonstrate two differences between \emph{HJQE} and the previous TER-based dataset.
2) To make use of the large scale translation parallel corpus, we propose self-supervised pre-training approach with two automatic tag correcting strategies to make the TER-based artificial dataset more consistent with human judgment and then boost the performance by large-scale pre-training.
3) We conduct experiments on our collected \emph{HJQE} dataset as well as the TER-based dataset MLQE-PE. Experimental results of the automatic and human evaluation show that our approach achieves higher consistency with human judgment.

\begin{table*}[]
\centering
\resizebox{.95\textwidth}{!}{
\begin{tabular}{lccccclcccc}
\toprule
\multirow{2}{*}{\textbf{Dataset}}   & \multirow{2}{*}{\textbf{Split}} & \multicolumn{4}{c}{\textbf{English-German}}          &  & \multicolumn{4}{c}{\textbf{English-Chinese}}         \\ \cmidrule{3-6} \cmidrule{8-11} 
                                    &                                 & samples & tokens & MT BAD tags     & MT Gap BAD tags &  & samples & tokens & MT BAD tags     & MT Gap BAD tags \\ \midrule
\multirow{2}{*}{\textbf{MLQE-PE}} & train                           & 7000    & 112342 & 31621 (28.15\%) & 5483 (4.59\%)   &  & 7000    & 120015 & 65204 (54.33\%) & 10206 (8.04\%)  \\
                                    & valid                           & 1000    & 16160  & 4445 (27.51\%)  & 716 (4.17\%)    &  & 1000    & 17063  & 9022 (52.87\%)  & 1157 (6.41\%)   \\ \midrule
\multirow{3}{*}{\textbf{\emph{HJQE} (ours)}} & train                           & 7000    & 112342 & 10804 (9.62\%)  & 640 (0.54\%)    &  & 7000    & 120015 & 19952 (16.62\%) & 348 (0.27\%)    \\
                                    & valid                           & 1000    & 16160  & 1375 (8.51\%)   & 30 (0.17\%)     &  & 1000    & 17063  & 2459 (14.41\%)  & 8 (0.04\%)      \\
                                    & test                            & 1000    & 16154  & 993 (6.15\%)    & 28 (0.16\%)     &  & 1000    & 17230  & 2784 (16.16\%)  & 11 (0.06\%)     \\ \bottomrule
\end{tabular}}
\caption{Statistics of TER-based MLQE-PE dataset and our proposed \emph{HJQE} dataset.}
\label{tab:data_statistics}
\end{table*}


\begin{figure}
    \centering
    \resizebox{.46\textwidth}{!}{
    \includegraphics{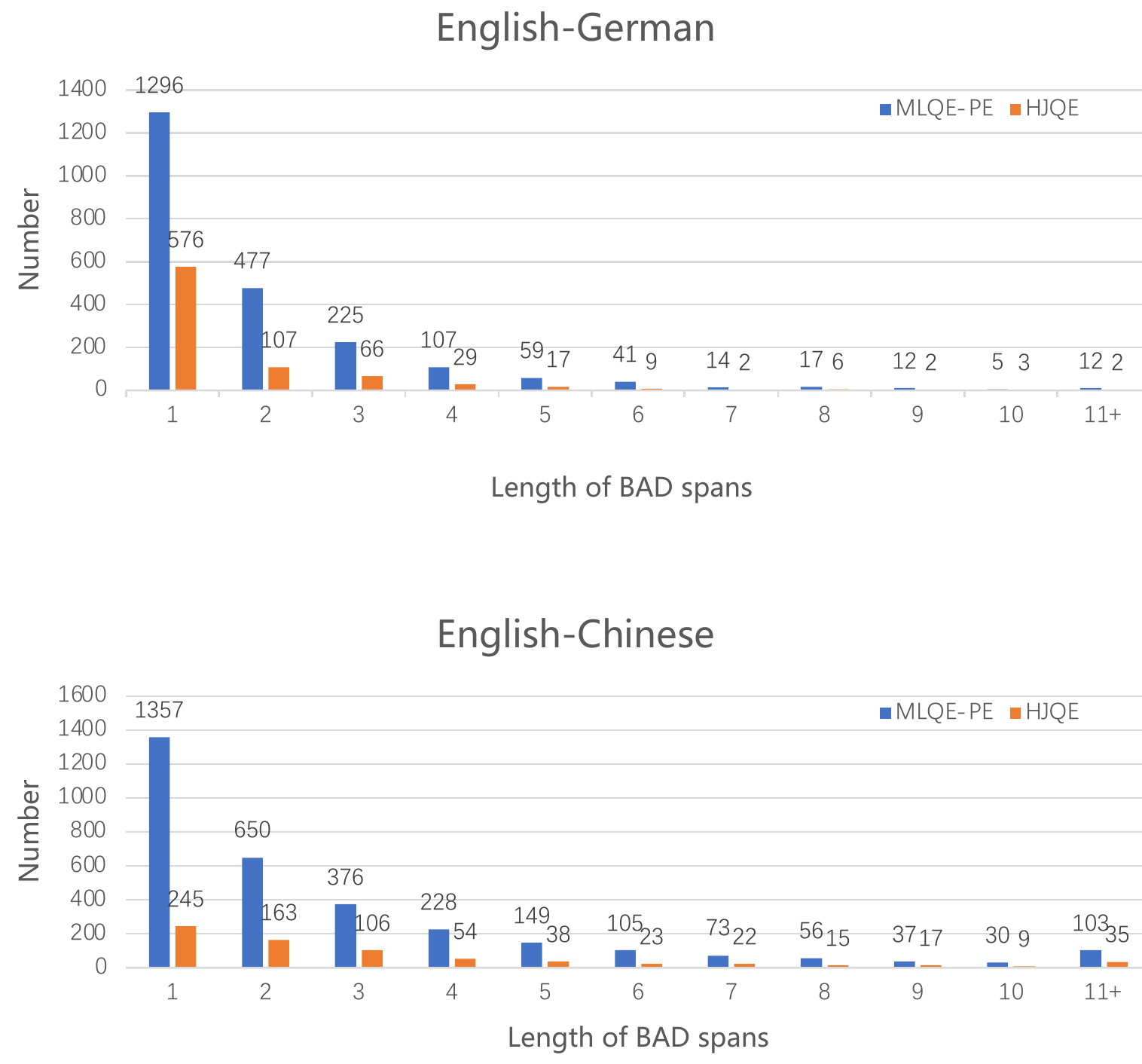}}
    \caption{The length distribution of BAD spans on English-German and English-Chinese validation sets. }
    \label{fig:length_of_bad_spans}
    \vspace{-0.2cm}
\end{figure}

\section{Data Collection and Analysis}
\label{data_collection_and_analysis}

\subsection{Data Collection}
To make our collected dataset comparable to TER-generated ones, we directly take the source and MT texts from MLQE-PE \cite{fomicheva2020mlqe}, the official dataset for the WMT20 QE shared task. It includes two language pairs that contain TER-generated annotations: English-German (En-De) and English-Chinese (En-Zh). The source texts are sampled from Wikipedia documents and the translations are obtained from the Transformer-based neural machine translation (NMT) system \cite{vaswani2017attention}.

Our data collection follows the following process. First, we hire a number of translator experts, where 5 translators for En-Zh and 6 for En-De. They are all graduated students that major in the translation and have the professional ability on the corresponding translation direction. For each sample, we randomly distribute it to two annotators. Each annotator is provided only the source sentence and its corresponding translation (but without the context or passage which the source sentence is taken from). For En-Zh, the translations are tokenized (as they are in MLQE-PE). Note that although the PE sentences exist in MLQE-PE, the human annotators have no access to them, making the annotation process as fair and unbiased as possible. After one sample is both annotated by the two annotators, we check whether the annotations are consistent. If they have annotation conflicts, we will 
re-assign the sample to other two annotators until we get the consistent annotations. 

For the annotation protocol, we ask human translators to find words, phrases, clauses or even the whole sentences that contain translation error in MT sentences, and annotate them as \texttt{BAD} tags. Here, the translation error means the translation distorts the meaning of the source sentence, but excluding minor mismatches such as synonyms and punctuation. Meanwhile, if the translation does not conform to the grammar of the target language, they should also find them and annotate as \texttt{BAD}. The annotation and distribution of samples are automatically conducted through the annotation system. After all samples are annotated, we ask another translator (1 for En-Zh and 1 for En-De, and they do not participant in the annotation process), sampling a small proportion (400 samples) of the full annotated dataset and ensure the accuracy is above 98\%.

\subsection{Statistics and Analysis}


\begin{figure}
    \centering
    \resizebox{.47\textwidth}{!}{
    \includegraphics{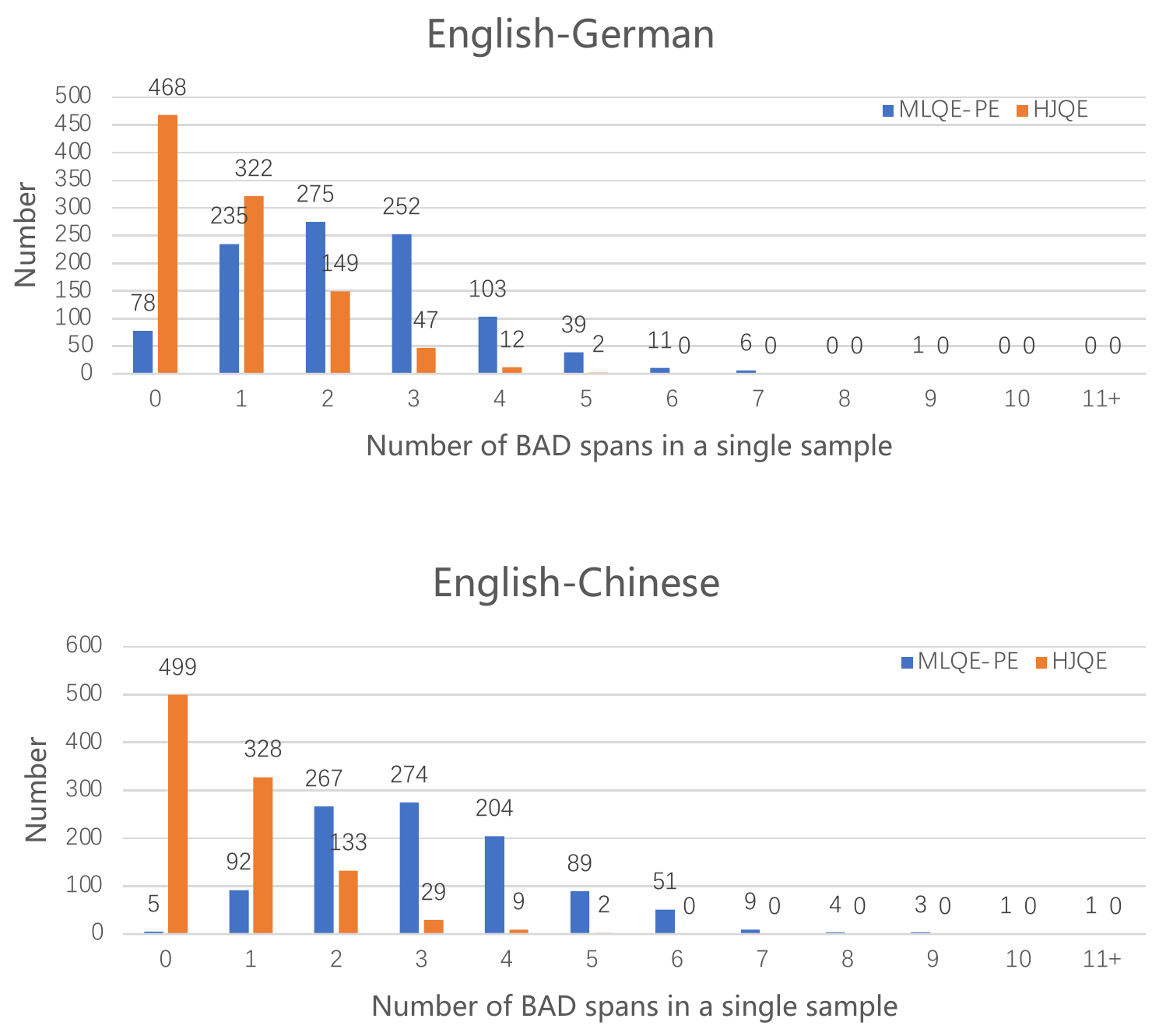}}
    \caption{The distribution that reveals how many \texttt{BAD} spans in every single validation sample.}
    \label{fig:coherence_of_bad_spans}
    \vspace{-0.3cm}
\end{figure}

\begin{figure*}
    \centering
    \resizebox{.9\textwidth}{!}{
    \includegraphics{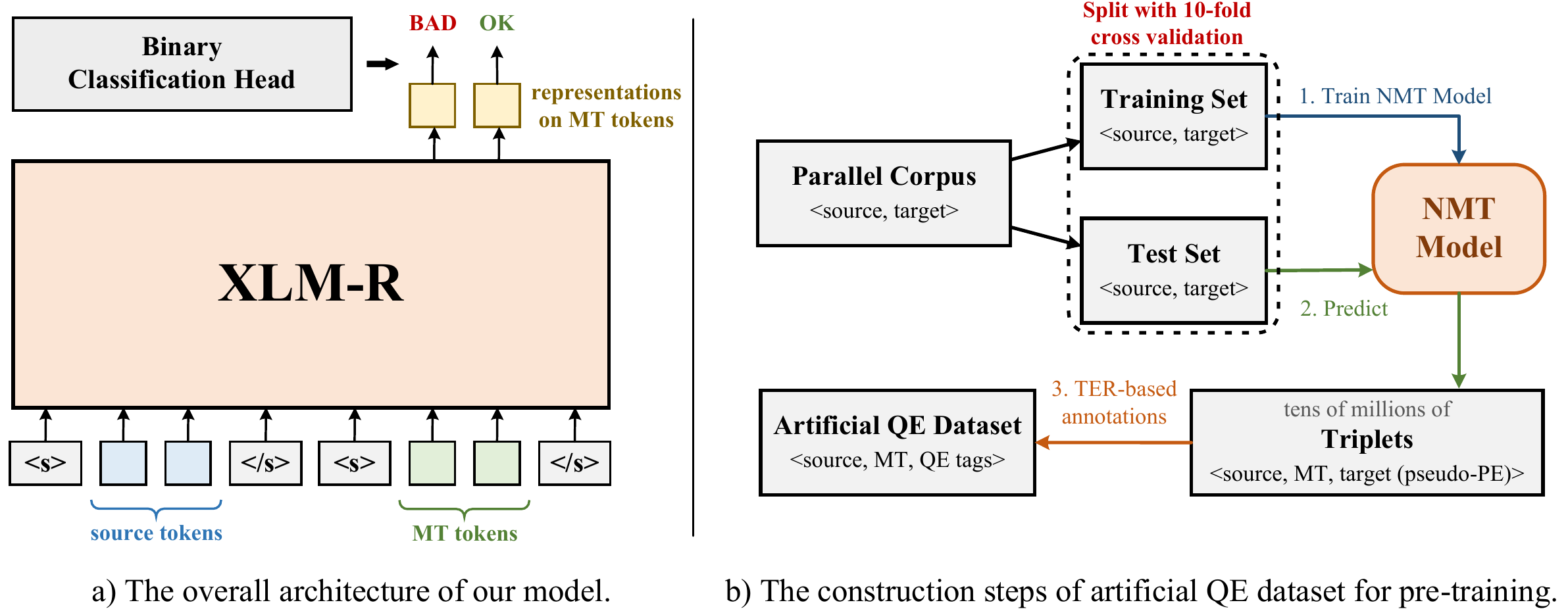}}
    \caption{The model architecture and the construction of artificial QE dataset.}
    \label{fig:model1}
    \vspace{-0.2cm}
\end{figure*}

\textbf{Overall Statistics.} In Table \ref{tab:data_statistics}, we show detailed statistics of the collected \emph{HJQE}. For comparison, we also present the statistics of MLQE-PE which is automatically annotated with TER. 
First, we see that the total number of \texttt{BAD} tags decreases heavily when human's annotations replaces the TER-based annotations (from 28.15\% to 9.62\% for En-De, and from 54.33\% to 16.62\% for En-Zh). It indicates that the human's annotations tends to annotate \texttt{OK} as long as the translation correctly expresses the meaning of the source sentence, but ignores the secondary issues like synonym substitutions and constituent reordering. 
Second, we find the number of \texttt{BAD} tags in the gap (indicating a few words are missing between two MT tokens) also greatly decreases. It's because that human's annotations tends to regard the missing translations (i.e., the \texttt{BAD} gaps) and the translation errors as a whole but only annotate \texttt{BAD} tags on MT tokens\footnote{As a result, we do not include the sub-task of predicting gap tags in our experiments.}.

\textbf{The Length of BAD Spans.} We show the number of \texttt{BAD} spans\footnote{Here, the \texttt{BAD} spans indicate the longest continuous tokens with \texttt{BAD} tags.} of different lengths in Figure \ref{fig:length_of_bad_spans}. We can see that most \texttt{BAD} spans only contain a few tokens, showing the well-known long-tail distribution. For En-De, the long-tail distribution is sharper, where 70.5\% of \texttt{BAD} spans are one-token spans. When comparing the TER-based annotations with human's, we find that human's annotation includes fewer \texttt{BAD} spans of each length, but the overall distribution is similar.

\textbf{Unity of BAD Spans.} To reveal the unity of the human's annotations, we group the samples according to the number of \texttt{BAD} spans in each single sample, and show the overall distribution. From Figure \ref{fig:coherence_of_bad_spans}, we can find that the TER-based annotations follow the Gaussian distribution, where a large proportion of samples contain 2, 3, or even more \texttt{BAD} spans, indicating the TER-based annotations are fragmented. However, our collected annotations on translation errors are more unified, with only a small proportion of samples including more than 2 \texttt{BAD} spans. Besides, we find a large number of samples that are fully annotated as \texttt{OK} in human's annotations. However, the number is extremely small for TER-based annotations (78 in English-German and 5 for English-Chinese). This shows a large proportion of \texttt{BAD} spans in TER-based annotations do not really destroy the semantic of translations and are thus regarded as \texttt{OK} by human annotators.


\section{Approach}


In this section, we will first introduce the backbone of the model and the self-supervised pre-training approach based on the large scale parallel corpus. Then, we propose two correcting strategies to make the TER-based artificial tags closer to the human judgment.

\begin{figure*}
    \centering
    \resizebox{\textwidth}{!}{
    \includegraphics{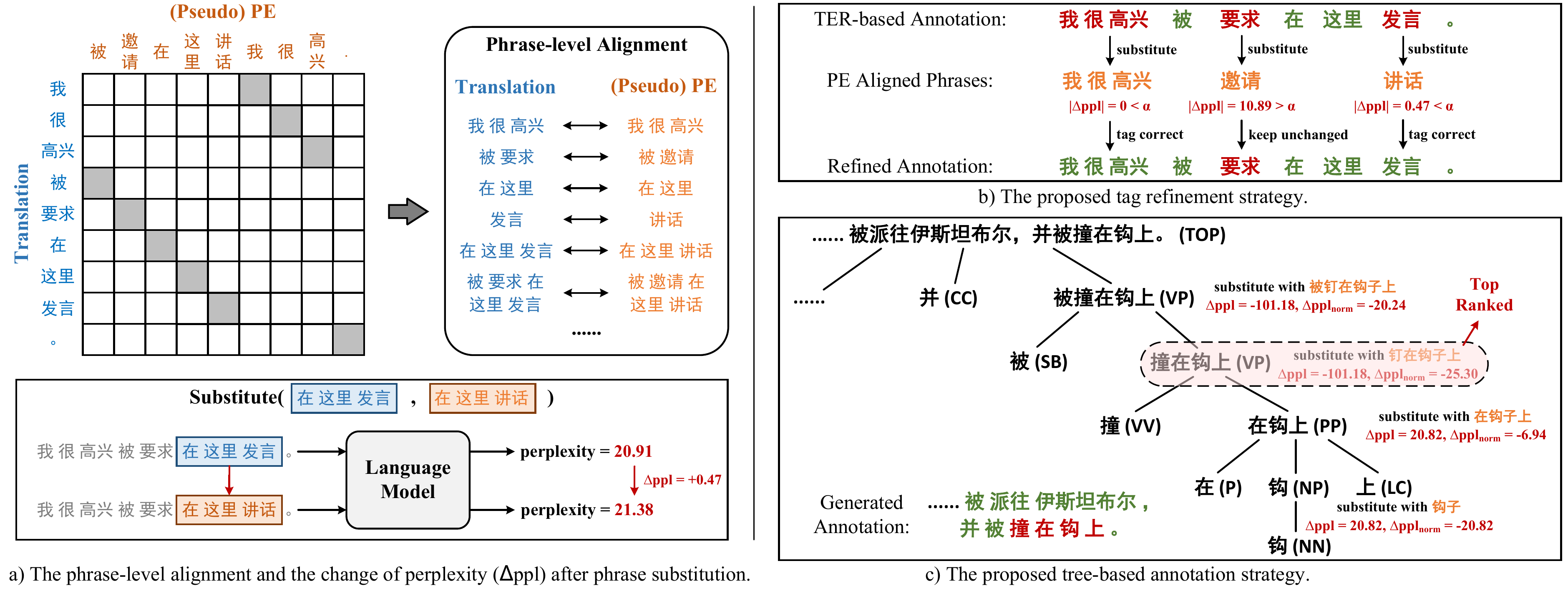}}
    \caption{The proposed two tag correcting strategies: Tag Refinement strategy and Tree-based Annotation strategy.}
    \label{fig:model2}
    \vspace{-0.2cm}
\end{figure*}

\subsection{Model Architecture} Following \cite{ranasinghe2020transquest, lee2020two, moura2020unbabel, ranasinghe2021exploratory}, we select the XLM-RoBERTa (XLM-R) \cite{conneau2020unsupervised} as the backbone of our model. XLM-R is a transformer-based masked language model pre-trained on large-scale multilingual corpus and demonstrates state-of-the-art performance on multiple cross-lingual downstream tasks.
As shown in Figure \ref{fig:model1}a, we concatenate the source sentence and the MT sentence together to make an input sample: $\boldsymbol{x}_i = \texttt{<s>} w^\text{src}_1, \dots, w^\text{src}_m \texttt{</s><s>} w^\text{mt}_1, \dots, w^\text{mt}_n \texttt{</s>}$, where $m$ is the length of the source sentence (src) and $n$ is the length of the MT sentence (mt). $\texttt{<s>}$ and $\texttt{</s>}$ are two special tokens to annotate the start and the end of the sentence in XLM-R, respectively.

For the $j$-th token $w^\text{mt}_j$ in the MT sentence, we take the corresponding representation from XLM-R for binary classification to determine whether $w_j$ belongs to good translation (\texttt{OK}) or contains translation error (\texttt{BAD}) and use the binary classification loss to train the model:
\begin{gather}
    s_{ij} = \sigma (\boldsymbol{w}^\mathsf{T} \text{XLM-R}_j(\boldsymbol{x}_i)) \label{eq:word_level_score} \\
    \mathcal{L}_{ij} = - (y \cdot \log s_{ij} + (1 - y) \cdot \log (1 - s_{ij}))
\end{gather}
where $\text{XLM-R}_j(\boldsymbol{x}_i) \in \mathbb{R}^d$ ($d$ is the hidden size of XLM-R) indicates the representation output by XLM-R corresponding to the token $w^\text{mt}_j$, $\sigma$ is the sigmoid function, $\boldsymbol{w} \in \mathbb{R}^{d \times 1}$ is the linear layer for binary classification and $y$ is the ground truth label.

\subsection{Self-Supervised Pre-training Approach} The translation knowledge contained in the parallel corpus of MT is very helpful for the QE task. As a result, we can adopt the parallel corpus to build artificial tags for self-supervised pre-training on QE. As shown in Figure \ref{fig:model1}b, the parallel corpus is firstly split into the training and the test set. Then the NMT model is trained with the training split and is used to generate translations for all sentences in the test split. From this, a large number of triplets are obtained, each consisting of source, MT, and target sentences. Finally, the target sentence is regarded as the pseudo-PE from the MT sentence, and the TER toolkit is used to generate word-level \texttt{OK|BAD} tags based on the principle of minimal editing (shown in the bottom of Figure \ref{fig:intro1}).

\subsection{Tag Correcting Strategies} As we discussed before, the two issues of TER-based tags limit the performance improvement of the self-supervised pre-training when applied to the downstream translation error detection task. In this section, we introduce two tag correcting strategies, namely tag refinement and tree-based annotation, that target these issues and make the TER-generated artificial QE tags more consistent with human judgment.

\textbf{Tag Refinement Strategy.}
In response to the first issue (i.e., wrong annotations due to the synonym substitution or constituent reordering), we propose the tag refinement strategy, which corrects the false \texttt{BAD} tags to \texttt{OK}.
Specifically, as shown in Figure \ref{fig:model2}a, we first generate the alignment between the MT sentence and the reference sentence (i.e., the pseudo-PE) using FastAlign\footnote{\url{https://github.com/clab/fast_align}} \cite{dyer2013simple}. Then we extract the phrase-to-phrase alignment through running the phrase extraction algorithm of NLTK\footnote{\url{https://github.com/nltk/nltk/blob/develop/nltk/translate/phrase_based.py}} \cite{bird2006nltk}. Once the phrase-level alignment is prepared, we substitute each \texttt{BAD} span with the corresponding aligned spans in the pseudo-PE and use the language model\footnote{\url{https://github.com/kpu/kenlm}} to calculate the change of the perplexity $\Delta ppl$ after this substitution.

If $| \Delta ppl | < \alpha$, where $\alpha$ is a hyper-parameter indicating the threshold, we regard that the substitution has little impact on the semantic and thus correct the \texttt{BAD} tags to \texttt{OK}. Otherwise, we regard the span does contain translation errors and keep the \texttt{BAD} tags unchanged (Figure \ref{fig:model2}b).

\begin{table*}[]
\centering
\resizebox{\textwidth}{!}{
\begin{tabular}{lcccclcccc}
\toprule
\multicolumn{1}{c}{\multirow{2}{*}{\textbf{Model}}} & \multicolumn{4}{c}{\textbf{English-German (En-De)}} &  & \multicolumn{4}{c}{\textbf{English-Chinese (En-Zh)}} \\ \cmidrule{2-5} \cmidrule{7-10} 
\multicolumn{1}{c}{}                                & \textbf{MCC}     & \textbf{F-OK}   & \textbf{F-BAD}   & \textbf{F-BAD-Span}  &  & \textbf{MCC}     & \textbf{F-OK}   & \textbf{F-BAD}   & \textbf{F-BAD-Span}   \\ \midrule
\multicolumn{10}{c}{\textit{Baselines}}                                                                                                                      \\
FT on \emph{HJQE} only                                         & 26.29   & \textbf{95.08}   & 31.09    & 20.97        &  & 38.56   & 90.76   & 47.56    & 26.66         \\
PT (TER-based)                                   & 9.52   & 34.62   & 13.54    & 3.09         &  & 15.17   & 36.66   & 31.53    & 2.40         \\
\ \ + FT on \emph{HJQE}                                        & 24.82   & 94.65   & 29.82    & 18.52        &  & 39.09   & \textbf{91.29}   & 47.04    & 25.93         \\ \midrule
\multicolumn{10}{c}{\textit{Pre-training only with tag correcting strategies (ours)}}                 \\
PT w/ Tag Refinement                              & 10.12*   & 49.33   & 14.32    & 3.62         &  & 19.36*   & 53.16   & 34.10    & 3.79          \\
PT w/ Tree-based Annotation                       & 8.94   & 84.50   & 15.84    & 6.94         &  & 21.53*   & 59.21   & 35.54    & 6.32          \\ \midrule
\multicolumn{10}{c}{\textit{Pre-training with tag correcting strategies + fine-tuning on \emph{HJQE} (ours)}}                 \\
PT w/ Tag Refinement + FT                                        & 27.54*   & 94.21   & \textbf{35.25}    & 21.13        &  & 40.35*   & 90.88   & 49.33    & 25.60         \\
PT w/ Tree-based Annotation + FT                                        & \textbf{27.67*}   & 94.44   & 32.41    & \textbf{21.38}        &  & \textbf{41.33*}   & 91.22   & \textbf{49.82}    & \textbf{27.21}         \\ \bottomrule
\end{tabular}}
\caption{The word-level QE performance on the test set of \emph{HJQE} for two language pairs, En-De and En-Zh. PT indicates pre-training and FT indicates fine-tuning. Results are all reported by $\times$100. The numbers with * indicate the significant improvement over the corresponding baseline with p < 0.05 under t-test \cite{semenick1990tests}.}
\label{tab:main_results}
\vspace{-0.2cm}
\end{table*}

\textbf{Tree-based Annotation Strategy.} Human's direct annotation tends to annotate the \textit{smallest} constituent that causes fatal translation errors \textit{as a whole} (e.g., the whole words, phrases, clauses, etc.). However, TER-based annotations are often fragmented, with the whole mis-translations being split into multiple \texttt{BAD} spans because some stop-words are aligned and labeled as \texttt{OK}. Besides, the \texttt{BAD} spans are often not well-formed in linguistics (e.g., two adjacent words but are from two different phrases). 

To address this issue, we propose the constituent tree-based annotation strategy. It can be regarded as an enhanced version of the tag refinement strategy that gets rid of the TER-based annotation. As shown in Figure \ref{fig:model2}c, we first generate the constituent tree for the MT sentences.\footnote{\url{https://stanfordnlp.github.io/stanza}} Each internal node (i.e., the non-leaf node) in the constituent tree represents a well-formed phrase such as noun phrase (NP), verb phrase (VP), prepositional phrase (PP), etc. For each node, we substitute it with the corresponding aligned phrase in the pseudo-PE. Then we still use the change of the perplexity $\Delta ppl$ to indicate whether the substitution of this phrase improves the fluency of the whole translation.

To only annotate the smallest constituents that exactly contain translation errors, we normalize $\Delta ppl$ by the number of words in the phrase and use this value to sort all internal nodes in the constituent tree: $\Delta ppl_\text{norm} = \frac{\Delta ppl}{r - l + 1}$, where $l$ and $r$ indicates the left and right position of the phrase, respectively. The words of a constituent node are integrally labeled as \texttt{BAD} only if $|\Delta ppl_\text{norm}| < \beta$ as well as there is no overlap with nodes that are higher ranked. $\beta$ is a hyperparameter indicating the threshold.


\section{Experiments}


\textbf{Datasets.} To verify the effectiveness of our proposed self-supervised pre-training approach with tag correcting strategies on detecting translation errors, we conduct experiments on both \emph{HJQE} and MLQE-PE \cite{fomicheva2020mlqe} datasets. MLQE-PE is the official dataset used in the WMT20 QE shared task \cite{specia-etal-2020-findings-wmt}, and \emph{HJQE} is our collected dataset with word-level annotations of translation errors. Note that MLQE-PE and \emph{HJQE} share the same source and MT sentences, thus they have exactly the same number of samples. We show the detailed statistics in Table \ref{tab:data_statistics}.
For the pre-training, we use the parallel dataset provided in the WMT20 QE shared task to generate the artificial QE dataset.

\textbf{Baselines.} To confirm the effectiveness of our proposed self-supervised pre-training approach with tag correcting strategies, we mainly select two baselines for comparison. In the one, we do not use the pre-training, but only fine-tune XLM-R on the training set of \emph{HJQE}. In the other, we pre-train the model on the TER-based artificial QE dataset and then fine-tune it on the training set of \emph{HJQE}.

\textbf{Evaluation.} Following WMT20 QE shared task \cite{specia-etal-2020-findings-wmt}, we use Matthews Correlation Coefficient (MCC) as the main metric and also provide the F1 score (F) for \texttt{OK}, \texttt{BAD} and \texttt{BAD} spans.\footnote{Please refer to Appendix \ref{sec:implementation_details} for implementation details.}


\subsection{Main Results} The results are shown in Table \ref{tab:main_results}. We can observe that the TER-based pre-training only brings very limited performance gain or even degrade the performance when compared to the ``FT on \emph{HJQE} only'' setting (-1.47 for En-De and +0.53 for En-Zh). It suggests that the inconsistency between TER-based and human's annotations leads to the limited effect of pre-training. However, when applying the tag correcting strategies to the pre-training dataset, the improvement is much more significant (+2.85 for En-De and +2.24 for En-Zh), indicating that the tag correcting strategies mitigate such inconsistency, improving the effect of pre-training.
On the other hand, when only the pre-training is applied, the tag correcting strategies can also improve the performance. It shows our approach can also be applied to the unsupervised setting, where no human-annotated dataset is available for fine-tuning.

\textbf{Tag Refinement v.s. Tree-based Annotation.} When comparing two tag correcting strategies, we find the tree-based annotation strategy is generally superior to the tag refinement strategy, especially for En-Zh. The MCC improves from 19.36 to 21.53 under the \textit{pre-training only} setting and improves from 40.35 to 41.33 under the \textit{pre-training then fine-tuning} setting. This is probably because the tag refinement strategy still requires the TER-based annotation and fixes based on it, while the tree-based annotation strategy actively selects the well-formed constituents to apply phrase substitution and gets rid of the TER-based annotation.

\textbf{Span-level Metric.} Through the span-level metric (F-BAD-Span), we want to measure the unity and consistency of the model's prediction against human judgment. From Table \ref{tab:main_results}, we find our models with tag correcting strategies also show higher F1 score on \texttt{BAD} spans (from 26.66 to 27.21 for En-Zh), while TER-based pre-training even do harm to this metric (from 26.66 to 25.93 for En-Zh). This phenomenon also confirms the aforementioned fragmented issue of TER-based annotations, and our tag correcting strategies, instead, improve the span-level metric by alleviating this issue.

\begin{table}[]
\resizebox{.48\textwidth}{!}{
\begin{tabular}{lcccccc}
\toprule
\multicolumn{1}{c}{\multirow{2}{*}{\textbf{\begin{tabular}[c]{@{}c@{}}Evaluate on →\\ Fine-tune on ↓\end{tabular}}}} & \multicolumn{3}{c}{\textbf{MLQE-PE}}             &           & \multicolumn{2}{c}{\textbf{\emph{HJQE}}} \\ \cmidrule{2-4} \cmidrule{6-7} 
\multicolumn{1}{c}{}                                                                                                 & \textbf{MCC*}  & \textbf{MCC}   & \textbf{F-BAD} & \textbf{} & \textbf{MCC}    & \textbf{F-BAD}  \\ \midrule
WMT20's best                                                                                                         & 59.28          & -              & -              &           & -               & -               \\ \midrule
\multicolumn{7}{c}{\textit{No pre-training (fine-tuning only)}}                                                                                                                                                         \\
MLQE-PE                                                                                                              & \textbf{58.21} & \textbf{46.81} & \textbf{75.02} &           & 22.49           & 34.34           \\
\emph{HJQE}                                                                                                                 & 49.77          & 23.68          & 36.10          &           & \textbf{45.76}  & \textbf{53.77}  \\ \midrule
\multicolumn{7}{c}{\textit{TER-based pre-training}}                                                                                                                                                                     \\
w/o fine-tune                                                                                                        & 56.51          & 33.58          & 73.85          &           & 11.38           & 27.41           \\
MLQE-PE                                                                                                              & \textbf{61.85} & \textbf{53.25} & \textbf{78.69} &           & 21.93           & 33.75           \\
\emph{HJQE}                                                                                                                 & 41.39          & 29.19          & 42.97          &           & \textbf{47.34}  & \textbf{55.43}  \\ \midrule
\multicolumn{7}{c}{\textit{Pre-training with tag refinement}}                                                                                                                                                           \\
w/o fine-tune                                                                                                        & 55.03          & 28.89          & 70.73          &           & 18.83           & 31.39           \\
MLQE-PE                                                                                                              & \textbf{61.35} & \textbf{48.24} & \textbf{77.17} &           & 21.85           & 33.31           \\
\emph{HJQE}                                                                                                                 & 39.56          & 25.06          & 67.40          &           & \textbf{47.61}  & \textbf{55.22}  \\ \bottomrule
\multicolumn{7}{c}{\textit{Pre-training with tree-based annotation}}                                                                                                                                                    \\
w/o fine-tune                                                                                                        & 55.21          & 26.79          & 68.11          &           & 20.98           & 32.84           \\
MLQE-PE                                                                                                              & \textbf{60.92} & \textbf{48.58} & \textbf{76.18} &           & 22.34           & 34.13           \\
\emph{HJQE}                                                                                                                 & 40.30          & 26.22          & 39.50          &           & \textbf{48.14}  & \textbf{56.02}  \\ \hline
\end{tabular}}
\caption{Performance comparison for En-Zh with different fine-tuning and evaluation settings. Since the test labels of MLQE-PE are not publicly available, we report the results on the validation set of both datasets. MCC* indicates the MCC score considering both the target tokens and the target gaps.}
\label{tab:analysis1}
\vspace{-0.2cm}
\end{table}

\subsection{Analysis}
\label{sec:analysis}

\textbf{Comparison to results on MLQE-PE.} To demonstrate the difference between the MLQE-PE (TER-generated tags) and our \emph{HJQE} datasets, and analyze how the pre-training and fine-tuning influence the results on both datasets, we compare the performance of different models on MLQE-PE and \emph{HJQE} respectively. The results for En-Zh are shown in Table~\ref{tab:analysis1}.

When comparing results in each group, we find that fine-tuning on the training set identical to the evaluation set is necessary for achieving high performance. Otherwise, fine-tuning provides marginal improvement (e.g., fine-tuning on MLQE-PE and evaluating on \emph{HJQE}) or even degrades the performance (e.g., fine-tuning on \emph{HJQE} and evaluating on MLQE-PE). This reveals the difference in data distribution between \emph{HJQE} and MLQE-PE.
Besides, we note that our best model on MLQE-PE outperforms WMT20's best model (61.85 v.s. 59.28) using the same MCC* metric, showing the strength of our model, even under the TER-based setting.

On the other hand, we compare the performance gain of different pre-training strategies. When evaluating on MLQE-PE, the TER-based pre-training brings higher performance gain (+6.44) than pre-training with two proposed tag correcting strategies (+1.43 and +1.77). While when evaluating on \emph{HJQE}, the case is opposite, with the TER-based pre-training bringing lower performance gain (+1.58) than tag refinement (+1.85) and tree-based annotation (+2.38) strategies. In conclusion, the pre-training always brings performance gain, no matter evaluated on MLQE-PE or \emph{HJQE}. However, the optimal strategy depends on the consistency between the pre-training dataset and the downstream evaluation task.

\begin{table}[]
\centering
\resizebox{.41\textwidth}{!}{
\begin{tabular}{lccccc}
\toprule
\multirow{2}{*}{\textbf{Scores}} & \multicolumn{2}{c}{\textbf{En-De}} & \multicolumn{1}{l}{} & \multicolumn{2}{c}{\textbf{En-Zh}} \\ \cmidrule{2-3} \cmidrule{5-6} 
                                 & \textbf{TER}     & \textbf{Ours}     & \textbf{}            & \textbf{TER}     & \textbf{Ours}     \\ \midrule
1 (terrible)                     & 3                & 1               &                      & 5                & 0               \\
2 (bad)                          & 36               & 16              &                      & 34               & 6               \\
3 (neutral)                      & 34               & 20              &                      & 29               & 21              \\
4 (good)                         & 26               & 61              &                      & 24               & 59              \\
5 (excellent)                    & 1                & 2               &                      & 8                & 14              \\ \midrule
Average score:                   & 2.86             & 3.47            &                      & 2.96             & 3.81            \\ 
\% Ours $\ge$ TER: & \multicolumn{2}{c}{89\%}  &  \multicolumn{1}{l}{}  & \multicolumn{2}{c}{91\%} \\ \bottomrule
\end{tabular}}
\caption{The results of human evaluation. We select the best-performed model fine-tuned on MLQE-PE and \emph{HJQE} respectively.}
\label{tab:analysis3}
\vspace{-0.2cm}
\end{table}

\textbf{Human Evaluation.} To evaluate and compare the models pre-trained on TER-based tags and corrected tags more objectively, human evaluation is conducted for both models. For En-Zh and En-De, we randomly select 100 samples (the source and MT sentences) from the validation set and use two models to predict word-level \texttt{OK} or \texttt{BAD} tags for them. Then, we ask human translators to give a score for each prediction, between 1 and 5, where 1 indicates the predicted tags are fully wrong, and 5 indicates the tags are fully correct. 

Table \ref{tab:analysis3} shows the results. We can see that the model pre-trained on corrected tags (Ours) achieves higher human evaluation scores than that pre-trained on TER-based tags on average. For about 90\% of samples, the prediction of the model pre-trained on corrected dataset can outperform or tie with the prediction of the model pre-trained on TER-based dataset. The results of human evaluation show that \emph{HJQE} is more consistent with human judgement.



\section{Related Work}

Early approaches on QE, such as QuEst \cite{specia2013quest} and QuEst++ \cite{specia2015multi}, mainly pay attention to the feature engineering. They aggregate various features and feed them to the machine learning algorithms for classification or regression. \citet{kim2017predictor} first propose the neural-based QE approach, called Predictor-Estimator. They first pre-train an RNN-based predictor on the large-scale parallel corpus that predicts the target word given its context and the source sentence. Then, they extract the features from the pre-trained predictor and use them to train the estimator for the QE task. This model achieves the best performance on the WMT17 QE shard task. After that, many variants of Predictor-Estimator are proposed \cite{fan2019bilingual, moura2020unbabel, cui2021directqe}. Among them, Bilingual Expert \cite{fan2019bilingual} replaces RNN with multi-layer transformers as the architecture of the predictor, and proposes the 4-dimension mismatching feature for each token. It achieves the best performance on WMT18 QE shared task. The Unbabel team also releases an open-source framework for QE, called OpenKiwi \cite{kepler2019openkiwi}, that implements the most popular QE models with configurable architecture.

Recently, with the development of pre-trained language models, many works select the cross-lingual language model XLM-RoBERTa \cite{conneau2020unsupervised} as the backbone \cite{ranasinghe2020transquest, lee2020two, moura2020unbabel, rubino2020intermediate, ranasinghe2021exploratory, zhao2021verdi}. Many works also explore the joint learning or transfer learning of the multilingual QE task (i.e., on many language pairs) \cite{sun2020exploratory, ranasinghe2020transquest, ranasinghe2021exploratory}. Meanwhile, on the word-level QE, \citet{fomicheva2021eval4nlp} propose a shared task with the new-collected dataset on explainable QE, aiming to provide word-level hints for sentence-level QE score. \citet{freitag2021experts} also study multidimensional human evaluation for MT and collect a large-scale dataset.

The QE model can be applied to the Computer-Assisted Translation (CAT) system together with other models like translation suggestion (TS) or automatic post-edit (APE). \citet{wang2020computer} and \citet{lee2021intellicat} use the QE model to identify which parts of the machine translations need to be correct, and the TS \cite{yang2021wets} also needs the QE model to determine error spans before giving translation suggestions.

\section{Conclusion}
In this paper, we focus on the task of word-level QE in machine translation and target the inconsistency issues between the TER-based QE dataset and human judgment. We first collect and release a benchmark dataset called \emph{HJQE} that reflects the human judgement on the translation errors in MT sentences. Besides, we propose the self-supervised pre-training approach with two tag correcting strategies, which makes the TER-based annotations closer to the human judgement and improves the final performance on the proposed benchmark dataset \emph{HJQE}. We conduct thorough experiments and analyses, demonstrating the necessity of our proposed dataset and the effectiveness of our proposed approach. Our future directions include improving the performance of phrase-level alignment, introducing phrase-level semantic matching, and applying data augmentation. We hope our work will provide a new perspective for future researches on quality estimation.

\section*{Broader Impacts}

Quality estimation often serves as a post-processing module in recent commercial machine translation systems. It can be used to indicate the overall translation quality or detect the specific translation errors in the sentences. This work focuses on the direct annotation of translation errors, training the model to fit the human judgment at the word level. To do this, we collect a new QE dataset and propose tag correcting strategies to force the TER-based artificial dataset used in the pre-training phase closer to human judgment. When applying our approach, the users should pay special attention to the following: a) The data source of \emph{HJQE} is Wikipedia, so our model should perform well on a similar domain but may perform poorly on other irrelevant domains. b) Since our approach is still data-driven, the data (as well as the pre-training parallel dataset) should be ethical and unbiased, or unexpected problems may arise. c) The proposed tag correcting strategies work well on En-De and En-Zh, but do not necessarily applicable to other language pairs since the characteristics among target languages are different. d) Since the system is neural-based, the interpretability is limited. It can still mistakenly annotate some forbidden or sensitive words to \texttt{OK} and cause unexpected accidents.


\bibliography{anthology,custom}

\begin{thebibliography}{32}
\expandafter\ifx\csname natexlab\endcsname\relax\def\natexlab#1{#1}\fi

\bibitem[{Bird(2006)}]{bird2006nltk}
Steven Bird. 2006.
\newblock Nltk: the natural language toolkit.
\newblock In \emph{Proceedings of the COLING/ACL 2006 Interactive Presentation
  Sessions}, pages 69--72.

\bibitem[{Che et~al.(2010)Che, Li, and Liu}]{che2010ltp}
Wanxiang Che, Zhenghua Li, and Ting Liu. 2010.
\newblock Ltp: A chinese language technology platform.
\newblock In \emph{Coling 2010: Demonstrations}, pages 13--16.

\bibitem[{Conneau et~al.(2020)Conneau, Khandelwal, Goyal, Chaudhary, Wenzek,
  Guzm{\'a}n, Grave, Ott, Zettlemoyer, and Stoyanov}]{conneau2020unsupervised}
Alexis Conneau, Kartikay Khandelwal, Naman Goyal, Vishrav Chaudhary, Guillaume
  Wenzek, Francisco Guzm{\'a}n, {\'E}douard Grave, Myle Ott, Luke Zettlemoyer,
  and Veselin Stoyanov. 2020.
\newblock Unsupervised cross-lingual representation learning at scale.
\newblock In \emph{Proceedings of the 58th Annual Meeting of the Association
  for Computational Linguistics}, pages 8440--8451.

\bibitem[{Cui et~al.(2021)Cui, Huang, Li, Geng, Zheng, Huang, and
  Chen}]{cui2021directqe}
Qu~Cui, Shujian Huang, Jiahuan Li, Xiang Geng, Zaixiang Zheng, Guoping Huang,
  and Jiajun Chen. 2021.
\newblock Directqe: Direct pretraining for machine translation quality
  estimation.
\newblock In \emph{Proceedings of the AAAI Conference on Artificial
  Intelligence}, volume~35, pages 12719--12727.

\bibitem[{Dyer et~al.(2013)Dyer, Chahuneau, and Smith}]{dyer2013simple}
Chris Dyer, Victor Chahuneau, and Noah~A Smith. 2013.
\newblock A simple, fast, and effective reparameterization of ibm model 2.
\newblock In \emph{Proceedings of the 2013 Conference of the North American
  Chapter of the Association for Computational Linguistics: Human Language
  Technologies}, pages 644--648.

\bibitem[{Fan et~al.(2019)Fan, Wang, Li, Zhou, Chen, and Si}]{fan2019bilingual}
Kai Fan, Jiayi Wang, Bo~Li, Fengming Zhou, Boxing Chen, and Luo Si. 2019.
\newblock “bilingual expert” can find translation errors.
\newblock In \emph{Proceedings of the AAAI Conference on Artificial
  Intelligence}, volume~33, pages 6367--6374.

\bibitem[{Fomicheva et~al.(2021)Fomicheva, Lertvittayakumjorn, Zhao, Eger, and
  Gao}]{fomicheva2021eval4nlp}
Marina Fomicheva, Piyawat Lertvittayakumjorn, Wei Zhao, Steffen Eger, and Yang
  Gao. 2021.
\newblock The eval4nlp shared task on explainable quality estimation: Overview
  and results.
\newblock In \emph{Proceedings of the 2nd Workshop on Evaluation and Comparison
  of NLP Systems}, pages 165--178.

\bibitem[{Fomicheva et~al.(2020)Fomicheva, Sun, Fonseca, Blain, Chaudhary,
  Guzm{\'a}n, Lopatina, Specia, and Martins}]{fomicheva2020mlqe}
Marina Fomicheva, Shuo Sun, Erick Fonseca, Fr{\'e}d{\'e}ric Blain, Vishrav
  Chaudhary, Francisco Guzm{\'a}n, Nina Lopatina, Lucia Specia, and
  Andr{\'e}~FT Martins. 2020.
\newblock Mlqe-pe: A multilingual quality estimation and post-editing dataset.
\newblock \emph{arXiv preprint arXiv:2010.04480}.

\bibitem[{Freitag et~al.(2021)Freitag, Foster, Grangier, Ratnakar, Tan, and
  Macherey}]{freitag2021experts}
Markus Freitag, George Foster, David Grangier, Viresh Ratnakar, Qijun Tan, and
  Wolfgang Macherey. 2021.
\newblock Experts, errors, and context: A large-scale study of human evaluation
  for machine translation.
\newblock \emph{arXiv preprint arXiv:2104.14478}.

\bibitem[{Heafield(2011)}]{heafield2011kenlm}
Kenneth Heafield. 2011.
\newblock Kenlm: Faster and smaller language model queries.
\newblock In \emph{Proceedings of the sixth workshop on statistical machine
  translation}, pages 187--197.

\bibitem[{Kepler et~al.(2019)Kepler, Tr{\'e}nous, Treviso, Vera, and
  Martins}]{kepler2019openkiwi}
Fabio Kepler, Jonay Tr{\'e}nous, Marcos Treviso, Miguel Vera, and Andr{\'e}~FT
  Martins. 2019.
\newblock Openkiwi: An open source framework for quality estimation.
\newblock In \emph{Proceedings of the 57th Annual Meeting of the Association
  for Computational Linguistics: System Demonstrations}, pages 117--122.

\bibitem[{Kim et~al.(2017)Kim, Jung, Kwon, Lee, and Na}]{kim2017predictor}
Hyun Kim, Hun-Young Jung, Hongseok Kwon, Jong-Hyeok Lee, and Seung-Hoon Na.
  2017.
\newblock Predictor-estimator: Neural quality estimation based on target word
  prediction for machine translation.
\newblock \emph{ACM Transactions on Asian and Low-Resource Language Information
  Processing (TALLIP)}, 17(1):1--22.

\bibitem[{Kingma and Ba(2014)}]{kingma2014adam}
Diederik~P Kingma and Jimmy Ba. 2014.
\newblock Adam: A method for stochastic optimization.
\newblock \emph{arXiv preprint arXiv:1412.6980}.

\bibitem[{Lee(2020)}]{lee2020two}
Dongjun Lee. 2020.
\newblock Two-phase cross-lingual language model fine-tuning for machine
  translation quality estimation.
\newblock In \emph{Proceedings of the Fifth Conference on Machine Translation},
  pages 1024--1028.

\bibitem[{Lee et~al.(2021)Lee, Ahn, Park, and Jo}]{lee2021intellicat}
Dongjun Lee, Junhyeong Ahn, Heesoo Park, and Jaemin Jo. 2021.
\newblock Intellicat: Intelligent machine translation post-editing with quality
  estimation and translation suggestion.
\newblock \emph{arXiv preprint arXiv:2105.12172}.

\bibitem[{Moura et~al.(2020)Moura, Vera, van Stigt, Kepler, and
  Martins}]{moura2020unbabel}
Joao Moura, Miguel Vera, Daan van Stigt, Fabio Kepler, and Andr{\'e}~FT
  Martins. 2020.
\newblock Ist-unbabel participation in the wmt20 quality estimation shared
  task.
\newblock In \emph{Proceedings of the Fifth Conference on Machine Translation},
  pages 1029--1036.

\bibitem[{Qi et~al.(2020)Qi, Zhang, Zhang, Bolton, and Manning}]{qi2020stanza}
Peng Qi, Yuhao Zhang, Yuhui Zhang, Jason Bolton, and Christopher~D Manning.
  2020.
\newblock Stanza: A python natural language processing toolkit for many human
  languages.
\newblock \emph{arXiv preprint arXiv:2003.07082}.

\bibitem[{Ranasinghe et~al.(2020)Ranasinghe, Orasan, and
  Mitkov}]{ranasinghe2020transquest}
Tharindu Ranasinghe, Constantin Orasan, and Ruslan Mitkov. 2020.
\newblock Transquest: Translation quality estimation with cross-lingual
  transformers.
\newblock In \emph{Proceedings of the 28th International Conference on
  Computational Linguistics}, pages 5070--5081.

\bibitem[{Ranasinghe et~al.(2021)Ranasinghe, Orasan, and
  Mitkov}]{ranasinghe2021exploratory}
Tharindu Ranasinghe, Constantin Orasan, and Ruslan Mitkov. 2021.
\newblock An exploratory analysis of multilingual word-level quality estimation
  with cross-lingual transformers.
\newblock \emph{arXiv preprint arXiv:2106.00143}.

\bibitem[{Rubino and Sumita(2020)}]{rubino2020intermediate}
Raphael Rubino and Eiichiro Sumita. 2020.
\newblock Intermediate self-supervised learning for machine translation quality
  estimation.
\newblock In \emph{Proceedings of the 28th International Conference on
  Computational Linguistics}, pages 4355--4360.

\bibitem[{Semenick(1990)}]{semenick1990tests}
Doug Semenick. 1990.
\newblock Tests and measurements: The t-test.
\newblock \emph{Strength \& Conditioning Journal}, 12(1):36--37.

\bibitem[{Snover et~al.(2006)Snover, Dorr, Schwartz, Micciulla, and
  Makhoul}]{snover2006study}
Matthew Snover, Bonnie Dorr, Richard Schwartz, Linnea Micciulla, and John
  Makhoul. 2006.
\newblock A study of translation edit rate with targeted human annotation.
\newblock In \emph{Proceedings of the 7th Conference of the Association for
  Machine Translation in the Americas: Technical Papers}, pages 223--231.

\bibitem[{Specia et~al.(2020)Specia, Blain, Fomicheva, Fonseca, Chaudhary,
  Guzm{\'a}n, and Martins}]{specia-etal-2020-findings-wmt}
Lucia Specia, Fr{\'e}d{\'e}ric Blain, Marina Fomicheva, Erick Fonseca, Vishrav
  Chaudhary, Francisco Guzm{\'a}n, and Andr{\'e} F.~T. Martins. 2020.
\newblock \href {https://aclanthology.org/2020.wmt-1.79} {Findings of the {WMT}
  2020 shared task on quality estimation}.
\newblock In \emph{Proceedings of the Fifth Conference on Machine Translation},
  pages 743--764, Online. Association for Computational Linguistics.

\bibitem[{Specia et~al.(2015)Specia, Paetzold, and Scarton}]{specia2015multi}
Lucia Specia, Gustavo Paetzold, and Carolina Scarton. 2015.
\newblock Multi-level translation quality prediction with quest++.
\newblock In \emph{Proceedings of ACL-IJCNLP 2015 System Demonstrations}, pages
  115--120.

\bibitem[{Specia et~al.(2013)Specia, Shah, De~Souza, and
  Cohn}]{specia2013quest}
Lucia Specia, Kashif Shah, Jos{\'e}~GC De~Souza, and Trevor Cohn. 2013.
\newblock Quest-a translation quality estimation framework.
\newblock In \emph{Proceedings of the 51st Annual Meeting of the Association
  for Computational Linguistics: System Demonstrations}, pages 79--84.

\bibitem[{Sun et~al.(2020)Sun, Fomicheva, Blain, Chaudhary, El-Kishky,
  Renduchintala, Guzm{\'a}n, and Specia}]{sun2020exploratory}
Shuo Sun, Marina Fomicheva, Fr{\'e}d{\'e}ric Blain, Vishrav Chaudhary, Ahmed
  El-Kishky, Adithya Renduchintala, Francisco Guzm{\'a}n, and Lucia Specia.
  2020.
\newblock An exploratory study on multilingual quality estimation.
\newblock In \emph{Proceedings of the 1st Conference of the Asia-Pacific
  Chapter of the Association for Computational Linguistics and the 10th
  International Joint Conference on Natural Language Processing}, pages
  366--377.

\bibitem[{Tuan et~al.(2021)Tuan, El-Kishky, Renduchintala, Chaudhary,
  Guzm{\'a}n, and Specia}]{tuan-etal-2021-quality}
Yi-Lin Tuan, Ahmed El-Kishky, Adithya Renduchintala, Vishrav Chaudhary,
  Francisco Guzm{\'a}n, and Lucia Specia. 2021.
\newblock \href {https://aclanthology.org/2021.eacl-main.50} {Quality
  estimation without human-labeled data}.
\newblock In \emph{Proceedings of the 16th Conference of the European Chapter
  of the Association for Computational Linguistics: Main Volume}, pages
  619--625, Online. Association for Computational Linguistics.

\bibitem[{Vaswani et~al.(2017)Vaswani, Shazeer, Parmar, Uszkoreit, Jones,
  Gomez, Kaiser, and Polosukhin}]{vaswani2017attention}
Ashish Vaswani, Noam Shazeer, Niki Parmar, Jakob Uszkoreit, Llion Jones,
  Aidan~N Gomez, {\L}ukasz Kaiser, and Illia Polosukhin. 2017.
\newblock Attention is all you need.
\newblock In \emph{Advances in neural information processing systems}, pages
  5998--6008.

\bibitem[{Wang et~al.(2020)Wang, Wang, Ge, Shi, Zhao, and
  Fan}]{wang2020computer}
Ke~Wang, Jiayi Wang, Niyu Ge, Yangbin Shi, Yu~Zhao, and Kai Fan. 2020.
\newblock Computer assisted translation with neural quality estimation and
  auotmatic post-editing.
\newblock In \emph{Proceedings of the 2020 Conference on Empirical Methods in
  Natural Language Processing: Findings}, pages 2175--2186.

\bibitem[{Wang et~al.(2021)Wang, Thompson, and Iyyer}]{wang2021phrase}
Shufan Wang, Laure Thompson, and Mohit Iyyer. 2021.
\newblock Phrase-bert: Improved phrase embeddings from bert with an application
  to corpus exploration.
\newblock In \emph{Proceedings of the 2021 Conference on Empirical Methods in
  Natural Language Processing}, pages 10837--10851.

\bibitem[{Yang et~al.(2021)Yang, Zhang, Li, Meng, and Zhou}]{yang2021wets}
Zhen Yang, Yingxue Zhang, Ernan Li, Fandong Meng, and Jie Zhou. 2021.
\newblock Wets: A benchmark for translation suggestion.
\newblock \emph{arXiv preprint arXiv:2110.05151}.

\bibitem[{Zhao et~al.(2021)Zhao, Wu, Niu, Wang, and Wang}]{zhao2021verdi}
Mingjun Zhao, Haijiang Wu, Di~Niu, Zixuan Wang, and Xiaoli Wang. 2021.
\newblock Verdi: Quality estimation and error detection for bilingual corpora.
\newblock In \emph{Proceedings of the Web Conference 2021}, pages 3023--3031.

\end{thebibliography}
\bibliographystyle{acl_natbib}

\newpage

\begin{table*}[t]
\centering
\resizebox{\textwidth}{!}{
\begin{tabular}{lcccclcccc}
\toprule
\multicolumn{1}{c}{\multirow{2}{*}{\textbf{Model}}} & \multicolumn{4}{c}{\textbf{English-German (En-De)}} &  & \multicolumn{4}{c}{\textbf{English-Chinese (En-Zh)}} \\ \cmidrule{2-5} \cmidrule{7-10} 
\multicolumn{1}{c}{}                                & \textbf{MCC}     & \textbf{F-OK}   & \textbf{F-BAD}   & \textbf{F-BAD-Span}  &  & \textbf{MCC}     & \textbf{F-OK}   & \textbf{F-BAD}   & \textbf{F-BAD-Span}   \\ \midrule
\multicolumn{10}{c}{\textit{Baselines}}                                                                                                                      \\
FT on \emph{HJQE} only                                         & 34.69   & 94.28   & 40.38    & 28.65        &  & 45.76   & 91.96   & 53.77    & \textbf{29.84}         \\
PT (TER-based)                                   & 13.13   & 37.30   & 18.80    & 4.72         &  & 11.38   & 25.91   & 27.41    & 2.16         \\
\ \ + FT on \emph{HJQE}                                        & 35.02   & 94.00   & 40.86    & 26.68        &  & 47.34   & 91.30   & 55.43    & 28.53         \\ \midrule
\multicolumn{10}{c}{\textit{With tag correcting strategies (ours)}}                                                                                                      \\
PT w/ Tag Refinement                              & 13.26   & 52.43   & 19.78    & 6.42         &  & 18.83   & 53.29   & 31.39    & 3.48          \\
\ \ + FT on \emph{HJQE}                                        & 37.70   & 94.08   & 43.32    & 30.83        &  & 47.61   & \textbf{92.39}   & 55.22    & 28.33         \\
PT w/ Tree-based Annotation                       & 13.92   & 84.79   & 22.75    & 9.64         &  & 20.98   & 59.32   & 32.84    & 6.53          \\
\ \ + FT on \emph{HJQE}                                        & 37.03   & \textbf{94.46}   & 42.54    & 31.21        &  & 48.14   & 91.88   & 56.02    & 28.17         \\
PT w/ Both                              & 13.12   & 39.68   & 18.94    & 5.26         &  & 21.39   & 56.76   & 32.74    & 5.72          \\
\ \ + FT on \emph{HJQE}                                        & \textbf{38.90}   & 94.44   & \textbf{44.35}    & \textbf{32.21}        &  & \textbf{48.71}   & 90.74   & \textbf{56.47}    & 25.51         \\ \bottomrule
\end{tabular}}
\caption{The word-level QE performance on the validation set of \emph{HJQE} for two language pairs, En-De and En-Zh. PT indicates pre-training and FT indicates fine-tuning.}
\label{tab:main_results_on_valid}
\vspace{-0.5cm}
\end{table*}

\appendix

\section{Implementation Details}
\label{sec:implementation_details}

Our implementation of QE model is based on an open-source framework, OpenKiwi\footnote{\url{https://github.com/Unbabel/OpenKiwi}} \cite{kepler2019openkiwi}. We use the large-sized XLM-R model and obtain it from hugging-face's library\footnote{\url{https://huggingface.co/xlm-roberta-large}}. We use the KenLM\footnote{\url{https://kheafield.com/code/kenlm.tar.gz}} \cite{heafield2011kenlm} to train the language model on all target sentences in the parallel corpus and calculate the perplexity of the given sentence. For the tree-based annotation strategy, we obtain the constituent tree through LTP\footnote{\url{http://ltp.ai/index.html}} \cite{che2010ltp} for Chinese and through Stanza\footnote{\url{https://stanfordnlp.github.io/stanza/index.html}} \cite{qi2020stanza} for German. We set $\alpha$ to 1.0 and $\beta$ to -3.0 in our tag correcting strategies based on the case studies and empirical judgment. In the preprocessing phase, we filter out parallel samples that are too long or too short, and only reserve sentences with 10-100 tokens.

We pre-train the model on 8 NVIDIA Tesla V100 (32GB) GPUs for two epochs, with the batch size set to 8 for each GPU. Then we fine-tune the model on a single NVIDIA Tesla V100 (32GB) GPU for up to 10 epochs, with the batch size set to 8 as well. Early stopping is used in the fine-tuning phase, with the patience set to 20. We evaluate the model every 10\% steps in one epoch. The pre-training often takes more than 15 hours and the fine-tuning takes 1 or 2 hours. We use Adam \cite{kingma2014adam} to optimize the model with the learning rate set to 5e-6 in both the pre-training and fine-tuning phases. For all hyperparameters in our experiments, we manually tune them on the validation set of \emph{HJQE}.

\section{Main Results on the Validation Set} In Table \ref{tab:main_results_on_valid}, we also report the main results on the validation set of \emph{HJQE}.


\section{Case Study}

\begin{figure*}
    \centering
    \resizebox{\textwidth}{!}{
    \includegraphics{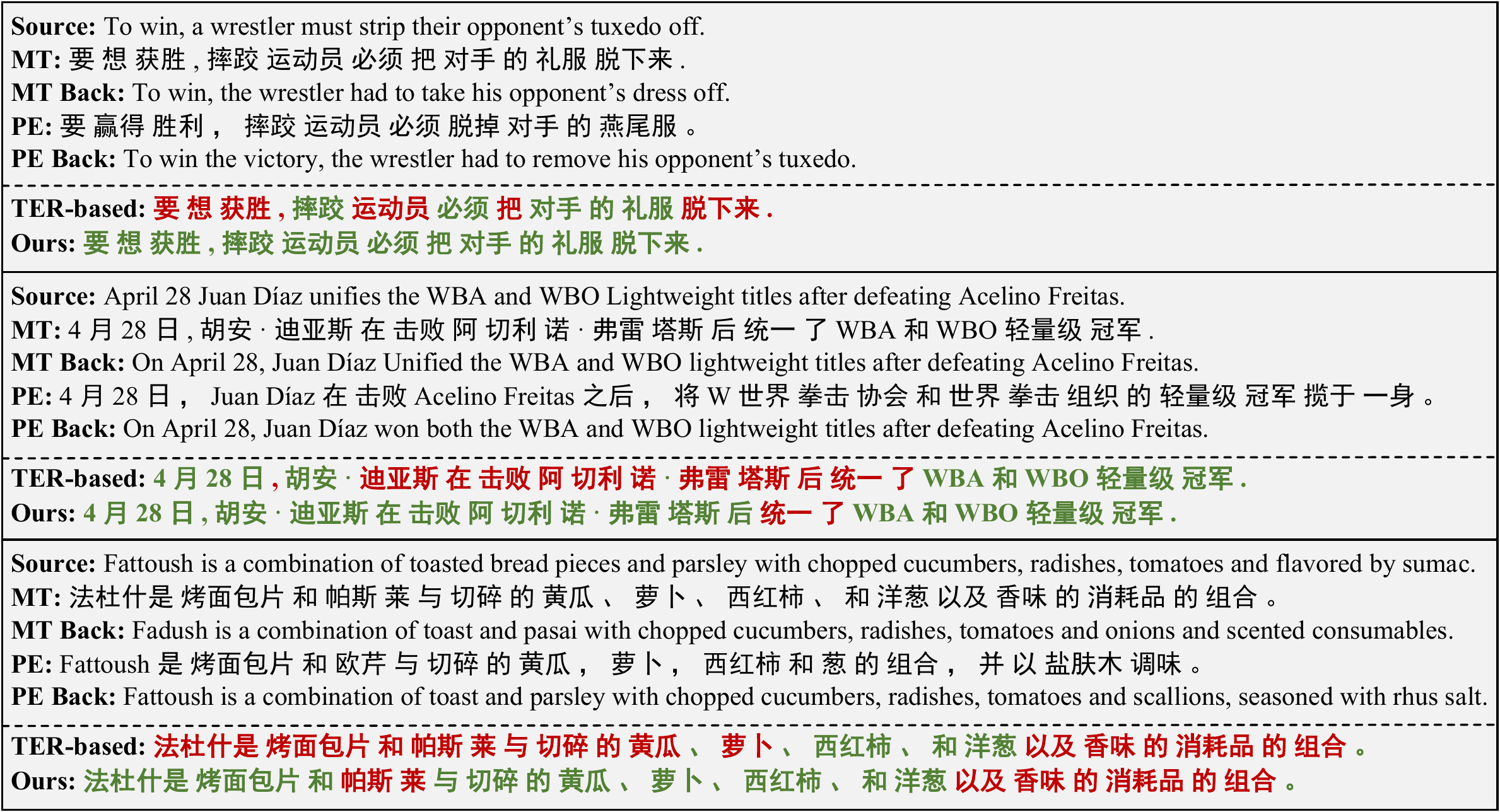}}
    \caption{Examples of word-level QE from the validation set of English-Chinese language pair.}
    \label{fig:good_cases}
    \vspace{-0.3cm}
\end{figure*}

In Figure \ref{fig:good_cases}, we show some cases from the validation set of English-Chinese language pair. From the examples, we can see that the TER-based model (noted as PE Effort Prediction) often annotates wrong \texttt{BAD} spans and is far from human judgment. For the first example, the MT sentence correctly reflects the meaning of the source sentence, and the PE is just a paraphrase of the MT sentence. Our model correctly annotates all words as \texttt{OK}, while TER-based one still annotates many \texttt{BAD} words. For the second example, the key issue is the translation of ``unifies'' in Chinese. Though ``统一'' is the direct translation of ``unifies'' in Chinese, it can not express the meaning of winning two titles in Chinese context. And our model precisely annotated the ``统一 了'' in the MT sentence as \texttt{BAD}. For the third example, the MT model fails to translate the ``parsley'' and the ``sumac'' to ``欧芹'' and ``盐肤木'' in Chinese, since they are very rare words. While the TER-based model mistakenly predicts long \texttt{BAD} spans, our model precisely identities both mistranslated parts in the MT sentence.

\section{Limitation and Discussion} We analyze some samples that are corrected by our tag correcting strategies and find a few bad cases. These are mainly because of the following: 1) There is noise from the parallel corpus (i.e., the source sentence and the target sentence are not well aligned). 2) The alignment generated by FastAlign contains unexpected errors, making some entries in the phrase-level alignments are missing or misaligned. 3) The scores given by KenLM (through the change of the perplexity after the phrase substitution) are sometimes not consistent with human judgment.

We also propose some possible solutions in response to the above problems as our future exploration direction. For the noise in the parallel corpus, we can use parallel corpus filtering methods that filter out samples with low confidence. We can also apply the data augmentation methods that expand the corpus based on the clean parallel corpus. For the errors by FastAlign, we may use a more accurate alignment model. For the scoring, we may introduce the neural-based phrase-level semantic matching model (e.g., Phrase-BERT \cite{wang2021phrase}) instead of the KenLM.

\end{CJK} 
\end{document}